\documentclass[final]{cvpr}

\usepackage{times}
\usepackage{epsfig}
\usepackage{graphicx}
\usepackage{amsmath}
\usepackage{amssymb}
\usepackage{makecell}

\usepackage[pagebackref=true,breaklinks=true,colorlinks,bookmarks=false]{hyperref}

\def\cvprPaperID{8331} 
\def\confYear{CVPR 2021}

\begin{document}

\title{Dogfight: Detecting Drones from Drones Videos}

\author{Muhammad Waseem Ashraf\textsuperscript{1}, Waqas Sultani\textsuperscript{1}, Mubarak Shah\textsuperscript{2}\\
\textsuperscript{1}Intelligent Machines Lab, Information Technology University, Pakistan\\
\textsuperscript{2}Center for Research in Computer Vision, University of Central Florida, USA\\
{\tt\small mohammadwaseem043@gmail.com, waqas.sultani@itu.edu.pk, shah@crcv.ucf.edu}
}

\maketitle
\thispagestyle{empty}
\pagestyle{empty}

\begin{abstract}
As airborne vehicles are becoming more autonomous and ubiquitous, it has become vital to develop the capability to detect the objects in their surroundings. This paper attempts to address the problem of drones detection from other flying drones. The erratic movement of the source and target drones, small size,  arbitrary shape, large intensity variations, and occlusion make this problem quite challenging. In this scenario, region-proposal based methods are not able to capture sufficient discriminative foreground-background information. Also, due to the extremely small size and complex motion of the source and target drones, feature aggregation based methods are unable to perform well. To handle this, instead of using region-proposal based methods, we propose to use a two-stage segmentation-based approach employing spatio-temporal attention cues. During the first stage, given the overlapping frame regions, detailed contextual information is captured over convolution feature maps using pyramid pooling. After that pixel and channel-wise attention is enforced on the feature maps to ensure accurate drone localization. In the second stage, first stage detections are verified and new probable drone locations are explored. To discover new drone locations, motion boundaries are used. This is followed by tracking candidate drone detections for a few frames, cuboid formation, extraction of the 3D convolution feature map, and drones detection within each cuboid. 
The proposed approach is evaluated on two publicly available drone detection datasets and outperforms several competitive baselines.

\end{abstract}
\begin{figure}[t]
    \centering
    \includegraphics[width=\linewidth]{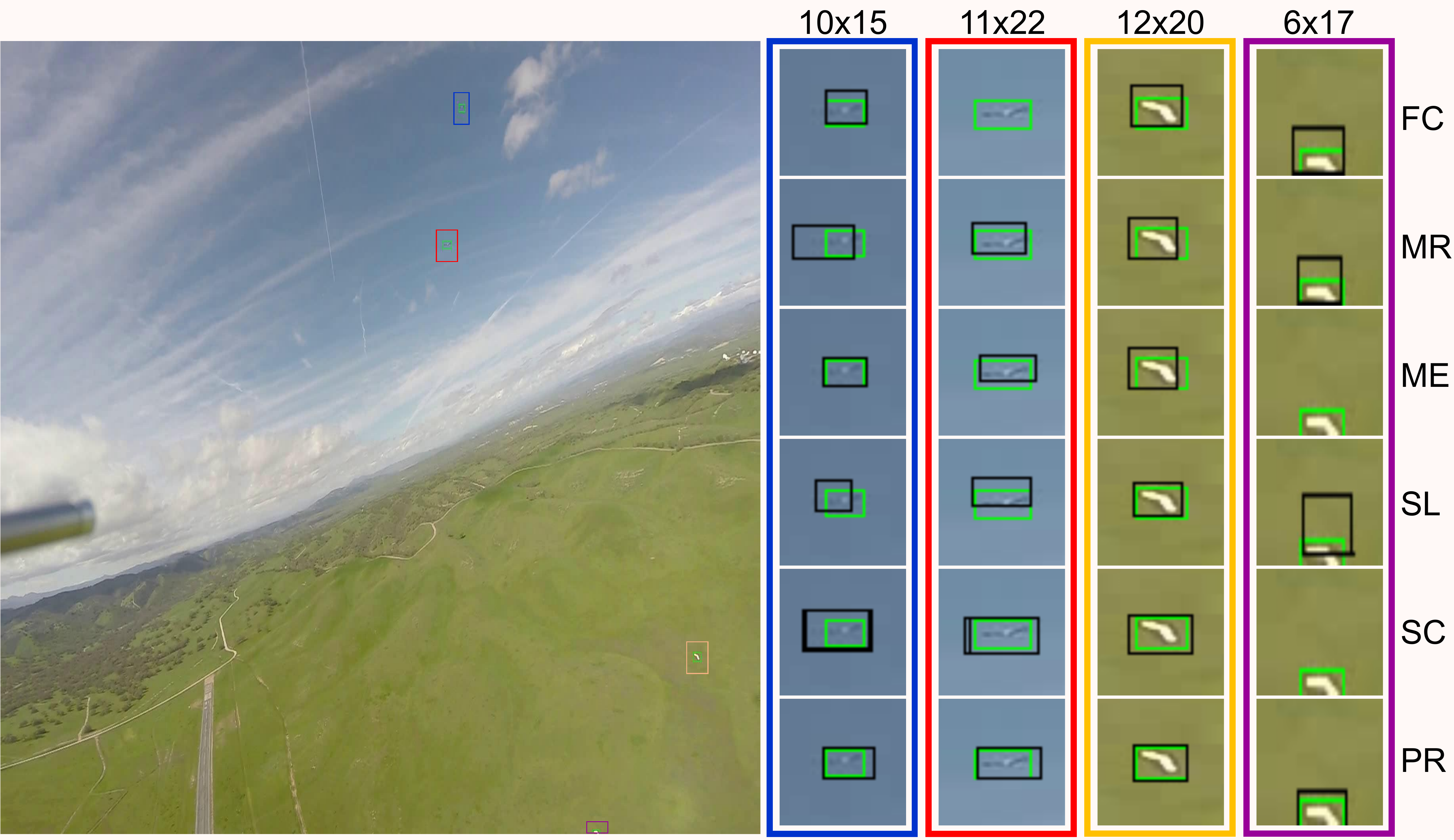}
    \caption{A comparison of our approach (PR) with state-of-the-art object detectors: FCOS (FC) \cite{tian2019fcos}, Mask-RCNN (MR) \cite{he2017mask}, MEGA (ME) \cite{chen2020memory}, SLSA (SL) \cite{wu2019sequence}, and  SCRDet (SC) \cite{yang2019scrdet}. In this frame (1080$\times$1920), there are four drones of sizes: 10$\times$15, 11$\times$22, 12$\times$20, 6$\times$17. The green bounding box represents the ground truth and output of detectors are shown in black colors.  For the clarity, we have not shown all false positives. The proposed approach provides better drone localization, reduces false positives and improves recall.} 
    \label{fig:method_comparison}
\end{figure}
\section{Introduction}

Drones are actively being used in several daily life applications such as agriculture \cite{UAV_Agriculture}, wildfire fighting \cite{UAV_WildfireFighing}, inventory applications \cite{UAV_Inventary_Application}, cinematography \cite{UAV_Cinema} and surveillance \cite{UAV_Survillance1,UAV_Survillance2}. Due to a large-scale application of drones, recently computer vision researchers have put forward several new techniques for object detection \cite{UAV_Object_Detection}, tracking \cite{UAV_Survillance2}, agriculture monitoring \cite{UAV_Agriculture} and human action recognition \cite{UCFARG,Okutama-Action} in the imagery obtained through drones. In addition to detecting different objects from a  drone video, it is also important to detect the drone itself from a video captured by another drone to avoid drone attacks \cite{UAV_Attack}, drone collisions \cite{UAV_PAMI} and safe multi-drone flights \cite{UAV_IROS, UAV_CVPR2015}.

Detection of ground objects and aerial drones from drone videos is a very challenging problem due to a large and abrupt camera motion, arbitrary drone shape and view changes, occlusion, and more importantly small object size. Although a lot of recent research has been conducted to detect and track ground objects and to detect human action using drones \cite{UAV_Object_Detection, UAV_Survillance1, UAV_Survillance2, UCFARG, Okutama-Action, sultani2020human}, limited work is being done to detect drones from drone videos \cite{UAV_IROS, UAV_PAMI}. 
To tackle the problem of drone detection, Li et al., \cite{UAV_IROS} proposed a new drone to drone detection dataset and employed handcrafted features for background estimation and foreground moving objects detection. Similarly, Rozantsev et al., \cite{UAV_PAMI} introduced a new challenging dataset of drones and air-crafts. They employed regression-based approaches to achieve object-centric stabilization and perform cuboid classification for detection purposes.

Usually flying drones occupy a few pixels in the video frames. For instance, the average drone size respectively is 0.05\%  and 0.07\% of the average frame size in drone detection datasets proposed in \cite{UAV_IROS} and  \cite{UAV_PAMI}. Note that this is much smaller than that of PASCAL VOC (22.62\%) and ImageNet (19.94\%).  Small objects including drones usually appear in the cluttered background and are oriented in different directions which makes its detection quite difficult. 
This issue was also pointed out by Huang et al., in \cite{small_object_low}, where they demonstrated that the mean Average Precision (mAP) of small objects is much lower than that of larger objects. Furthermore, the object detection performance further worsens in the videos \cite{chen2020memory}. 
To address this, Noh et al., \cite{noh2019better} proposed a feature-level super-resolution-based approach that utilizes high-resolution target features for supervising a low-resolution model. However, this would require the availability of both low and high-resolution drone images which are difficult to obtain in drone videos where the drone is already flying at a far distance.  Similarly,  Yang et al., \cite{yang2019scrdet} employed a region proposal-based multi-level feature fusion approach to detect small objects and introduce a new loss function to handle rotated objects. However, due to very small size objects, less salient, and cluttered backgrounds i.e., clouds, buildings, etc., it is difficult to obtain well-localized region proposals, specifically in drone detection datasets. Through adversarial learning, Wu et al., \cite{ObjectDetection_UAVs} proposed to learn domain-specific features employing metadata (flying altitudes, weather, and view angles). Given the recent low prices of drones,  it is more useful to use an RGB camera for detection and collision avoidance purposes instead of relying on the expensive hardware for metadata collection. Authors in \cite{zhu2017flow, liu2018mobile, wu2019sequence} proposed to use convolution feature aggregations across video frames to achieve improved video object detection. Our experimental results and analysis reveal that although feature aggregations \cite{zhu2017flow, liu2018mobile, wu2019sequence} techniques work well for large video object detection, for drone detection explicit motion information is more useful.

In this paper, we propose a two-stage segmentation-based approach to detect drones in cluttered backgrounds. The first stage uses only the appearance cues while the second stage exploits spatio-temporal cues. Given a video frame, we divide it into overlapping frame regions. Each frame region is passed through deep residual networks \cite{he2016deep} to obtain the convolution feature maps which are then followed by pyramid pooling layers \cite{zhao2017pyramid} to embed the contextual information. After that pixel-wise and channel-wise attention is employed on the convolution feature maps to discriminate drone boundaries from the background and achieve improved drone localization. The purpose of the second stage is to discover missing detections, remove false detections, and confirm the true positive detections by employing motion information. To discover the missing drones, we employ motion boundaries to find probable drone locations.
Given the detections from the first stage and motion boundaries locations, we track each location forward and backward for a few (eight) frames. After that, cuboids are extracted across those tracks and are fed to a 3D convolutional neural network \cite{carreira2017quo} for spatio-temporal feature extraction. This is followed by pyramid pooling layers. Similar to the first stage, we employ pixel and channel-wise attention in the second stage as well to get the improved localization.  The proposed approach significantly outperforms several competitive baselines. In the experimental section, we validate the efficacy of each step of the proposed approach. The rest of the paper is organized as follows. Section 2 provides a brief overview of the related developments in small object detection including drones in video and images. Section 3 deals with our proposed methodology and Section 4 covers experimental results. Finally, Section 5 concludes the paper.

\begin{figure*}[t]
    \centering
    \includegraphics[width=\linewidth, height=7cm]{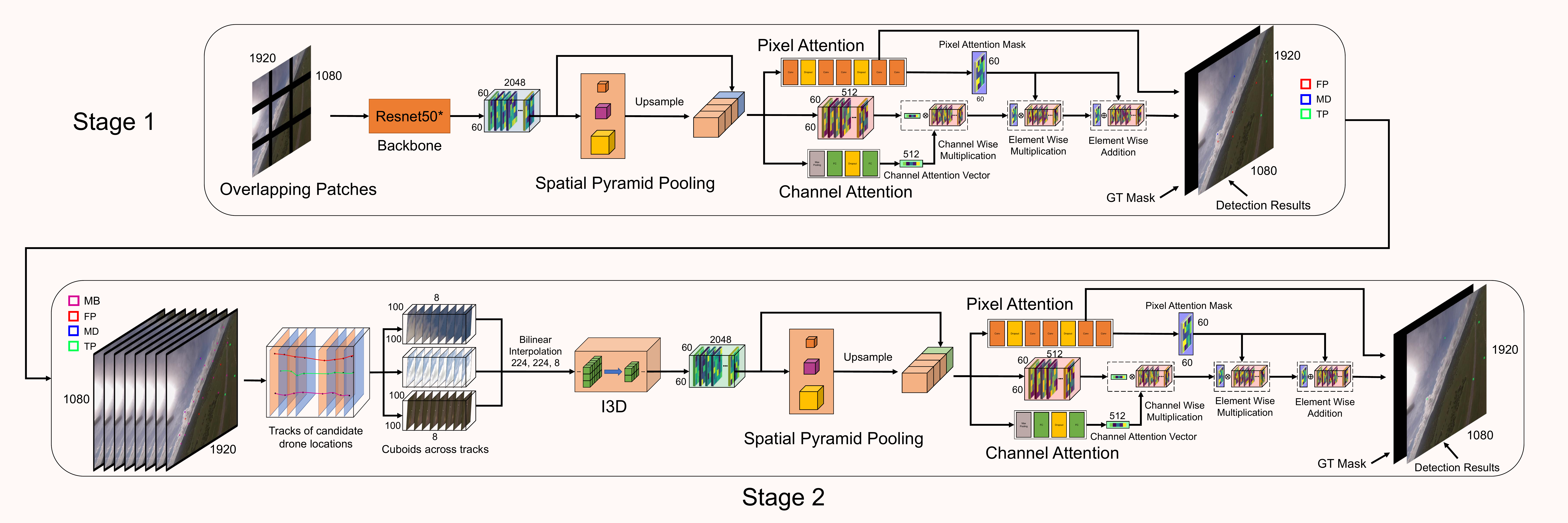}
    \caption{Our pipeline is divided into two stages. Stage-1 extracts Resnet50* features from the overlapping regions of each frame followed by pyramid pooling to retain global and local contextual information. Channel-wise and pixel-wise attention help in learning better localization of drones. Resnet50* refers to the modifications that we have applied (ref Section 3.1). Stage-2 combines spatial information with temporal data of the videos. Detections from  stage-1 along with candidate regions discovered using motion boundaries are used as candidate regions where UAV can exist. All the proposals are tracked for 8 frames in a forward and backward manner to generate cuboids of size  224$\times$224$\times$8. Each cuboid is passed through the I3D network followed by the attention network to accurately locate drones within each cuboid. In figure MD, TP, FP, and MB corresponds to missed detection, true positive, false positive, motion boundaries respectively.}
    \label{fig:pipeline}
\end{figure*}

\section{Related work}

\noindent\textbf{Object Detection:}
Recent years have witnessed tremendous
strides in improving object recognition accuracy \cite{he2017mask, redmon2018yolov3, ren2015faster, lin2017focal, cai2018cascade,law2018cornernet} on complex benchmark datasets. Law et al. \cite{law2018cornernet} proposed a single-stage network where they detect an object bounding box as a pair of key points. Also, the authors introduced a novel corner pooling scheme for better corners localization. To address the foreground-background class imbalance in object detection datasets, Lin et al. \cite{lin2017focal} presented a modified cross-entropy loss that reduces the weights of the
loss assigned to well-classified examples. Similarly, Li et al., \cite{Li_2020_CVPR} pointed out the underperformance of object detectors for  long-tail
distribution datasets. They proposed a novel balanced group-softmax to modulate the training process and ensure that the classifier is sufficiently trained for all the classes. Cai et al., \cite{cai2018cascade} put forward an idea of multi-stage object detection where a sequence of detectors trained with increasing
intersection over union (IoU) thresholds to achieve improved object localization. To further improve object detection accuracy, authors in \cite{guo2019augfpn} introduced a new feature fusion technique through multi-scale features summation in a feature pyramid. Although impressive detection results, most of these detectors have low detection and localization accuracy for small objects.

\noindent{\textbf{Small Object Detection:}}
To address the problem of small, cluttered, and oriented objects, Li et al., \cite{Li_2017_CVPR} proposed to use generative adversarial networks to decrease the gap of feature representation of small and large objects. To further improve feature representation for small objects, authors in \cite{noh2019better} proposed to jointly use the low and high-resolution feature maps of the same images during the training by matching the relative receptive fields between the pairs. Yang et al., \cite{yang2019scrdet} proposed a multi-layer feature fusion technique along with attention networks to achieve improved object detection for cluttered and rotated objects. However, most of these methods are region proposal based which are unable to capture sufficient foreground-background information specifically when the objects are extremely small (such as 0.05\% or 0.07\% of the average image size as in the case of drone detection).
A related problem to ours is the detection of ground objects from drone images such as \cite{Yang_2019_ICCV} which proposed to detect clustered objects in aerial images. However, while detecting drones from a drone, target objects (drones) undergo fast, abrupt, and highly unpredictable movements as compared to objects (car, human) on the ground. Target objects can have arbitrary changing shapes due to their rotation around any axis as compared to the restricted movements of ground objects. Furthermore, as compared to ground objects, drones have much smaller sizes, and the target objects appear/disappear more frequently due to clouds.

\noindent{\textbf{Video Object Detection:}}
Historically, taking a page from object detection, several video-based object detectors were introduced, including \cite{zhu2017flow,liu2018mobile,wu2019sequence,chen2020memory}.
The object detection accuracy degrades in the video due to several factors including motion blur, occlusion, and out of focus, etc,. To address this Zhu et al., \cite{zhu2017flow} proposed to include the feature representation of each frame by aggregation of nearby frames features using optical flow. Authors in \cite{liu2018mobile} improve feature aggregation speed using the Bottleneck-LSTM layer.  Instead of aggregating features from nearby frames, Wu et al., \cite{wu2019sequence} proposed to use whole sequence-level features aggregation for improved video object detection. Recently,  Chen et al., \cite{chen2020memory} utilized both the global semantic and local localization information employing memory aggregation network and have shown impressive results on several video object detection benchmarks. Although impressive results, most of these methods are tested on standard video object detection datasets where the object covers a significant portion of the video frame. Furthermore, motion is much more complex in drone videos as compared to standard video object detection datasets.

\noindent{\textbf{Drones Detection:}}
Recent low prices of drones will bring more and more UAVs to the sky. To keep the prices and weight of UAVs small, it is useful to detect other flying drones using simple RGB cameras instead of expensive radar systems. Therefore researchers have addressed the problem of drone detection for different applications. After obtaining a large number of spatio-temporal (s-t) tubes at different spatial resolutions, Rozantsev et al., \cite{UAV_PAMI} employs two CNN models to obtain coarse and fine motion stabilization in each s-t tube respectively. Next, they obtain UAVs detection by classifying each s-t tube using a third CNN.
There are several differences
between our method and \cite{UAV_PAMI}. 1) Instead of using computationally
expensive region-based sliding windows at several
scales, we employ an efficient fully convolutional
segmentation-based approach, 2) as compared to \cite{UAV_PAMI}, our
approach does not need perfectly drone-centered cuboids
and learn rich spatio-temporal information using I3D, and 3)
we employ attention networks to improve feature representation
for improved localization.
Similarly, authors in \cite{UAV_IROS, Deep_UAV} detected moving drones by subtracting background images and then identified UAVs using deeply learned classifiers. Furthermore, they use Kalman filtering to get improved detection. Their method uses a lot of parameters and thresholds which make it irreproducible.  Also, their reliance on background subtraction for moving object detection produces a lot of false positives. Instead of using RGB images, some researchers have also proposed to employ depth maps to achieve 3D localization \cite{carrio2018drone,vrba2019onboard}. However, depth maps are quite expensive to obtain in real-world scenarios and add the extra payload to UAVs.

\section{Proposed method}
Our goal is to detect and localize drones in the video frames which are captured by other flying drones.  Our proposed approach to tackle this challenging problem is based on the three observations: (1) due to very small drone size, region proposal based method may not be able to capture enough discriminative foreground-background information, therefore the bottom-up segmentation based approach which classifies each pixel is preferable; (2) the model should learn subtle visual differences between a drone and the background (clouds, etc); (3) due to large abrupt motion of target and source drones, the feature aggregation methods may not be sufficient and we need to use explicitly optical flow information as has been successfully employed in several action recognition works \cite{twostream_CVPR2018}. In the following, we first discuss details of the segmentation network (Sec. 3.1), followed by the attention networks used to get improved localization (Sec 3.2). Finally, we discuss how we use motion information to discover missing detection and thus improve recall.     

\subsection{Stage-1: Exploiting Spatial Cues}
We start with appearance-based pixel classification to accurately localize drones. For spatial feature computation, we resort to deep residual networks \cite{he2016deep}. However, given the extremely small size of drones (0.05\% or 0.07\% of image size), obtaining good discriminative features by utilizing the whole images is not possible. The standard 2D CNN networks such as Resnet50 require a fixed size input image (473$\times$473). Therefore image resizing from high-resolution image to low resolution (1080$\times$1920 to 473$\times$473) for feature computation can further decrease the spatial resolution of the drone to one or two pixels. Secondly, as the network goes deeper, we lose local information.
To address this, we use two steps: 1) To avoid resizing the image, we divide each frame into overlapping regions, 2) We modify Resnet50 to keep local information intact as we go deep in the network. Specifically, we extract features from all four blocks of Resnet50 \cite{he2016deep} and concatenate them together after spatial resizing of the first block to avoid dimension mismatch. Finally, we use 1$\times$1 convolution to get back the original dimension. We call modified Resnet50 as Resnet50*. Inspired by the use of pyramid pooling \cite{PSPNet} in several applications we employ pyramid pooling in our framework. Specifically, after obtaining features from Resnet50*, we apply the pyramid pooling using four different kernel sizes and concatenate those multi-scale features after up-sampling. 

In experiments, we observe that although the above network provides decent drone detection,  in several cases, it is unable to accurately detect and localize the drones.  Therefore, to make feature maps more focused on the foreground, we use pixel and channel-wise attentions. Pixel and channel-wise attention networks are described in the next section.

\subsection{Attention networks}
Assuming the drone of size 16$\times$11 (the average drone size in \cite{UAV_IROS}), by missing only a few pixels from both sides, the intersection over union (IOU) drops to below 0.5. Therefore, it is crucial to get accurate localization for true drone detection. To achieve this, we introduce detailed pixel-wise and channel-wise attention on the convolution feature maps. Recently, several attention networks \cite{NIPS2017_3f5ee243,SEC_Network,yang2019scrdet,Choe_2019_CVPR} have been introduced for different computer vision applications. 

\begin{figure}[ht]
    \centering
    \includegraphics[width=\linewidth]{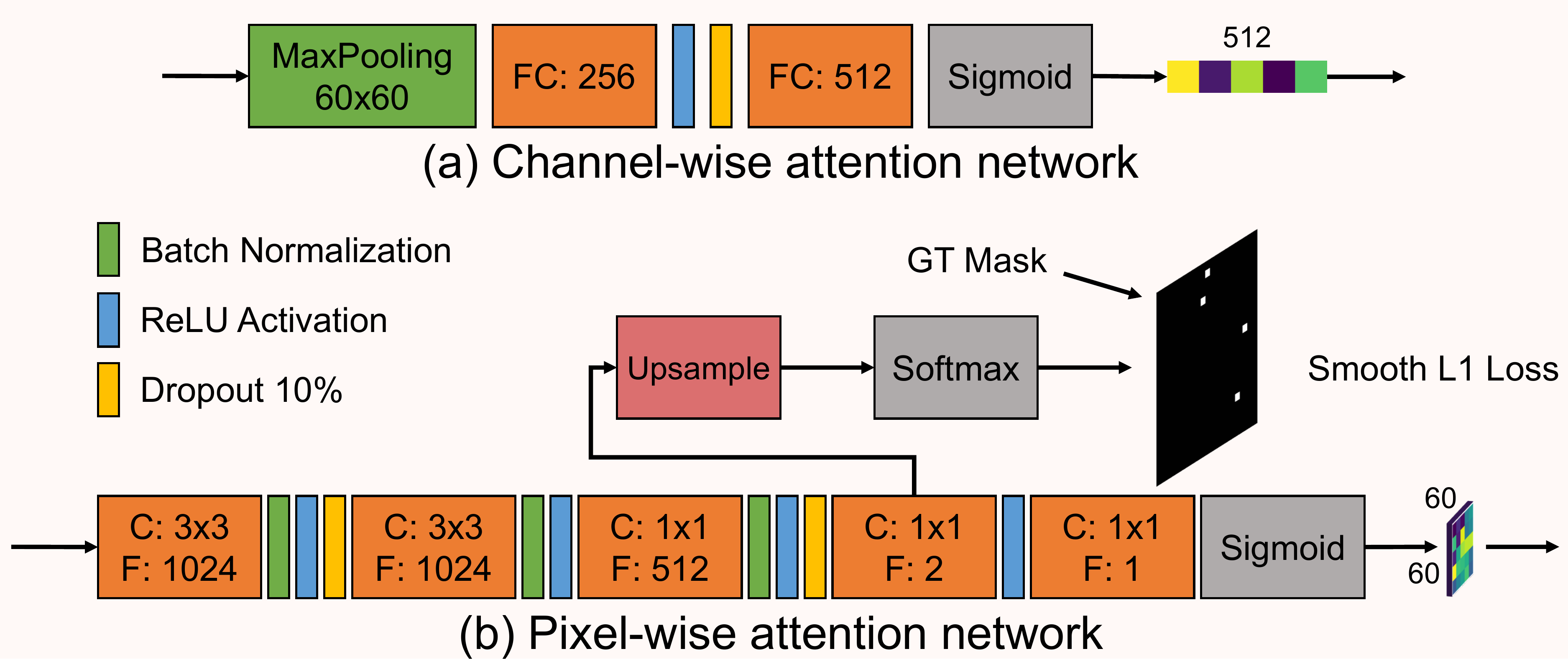}
    \caption{Architectural details of (a) channel-wise and (b) pixel-wise attention network, where `FC' represents fully connected layer with number of units in (a), `C' and `F' represent convolution and number of filters respectively in (b).}
    \label{fig:channel_pixel weighting_network}
\end{figure}

\noindent{\textbf{Channel-wise attention:}}
Inspired by \cite{SEC_Network}, we use a channel-wise attention network to automatically learn to give more weights to informative feature channels and suppress less informative ones. The architectural details of channel-wise attention networks are given in Figure \ref{fig:channel_pixel weighting_network} (a). This attention is achieved through channel-wise multiplication of attention vector with convolution feature maps.

\noindent{\textbf{Pixel-wise attention:}} Similar to channel-wise attention vector, we generate pixel-wise attention matrix to assign more weights to spatial location which corresponds to drones and less weight to non-drone regions similar to \cite{Choe_2019_CVPR,yang2019scrdet}. The architectural details of the pixel attention network are given in Figure \ref{fig:channel_pixel weighting_network} (b). To suppress background information, we perform element-wise multiplication of pixel attention mask with all convolution maps channels. This is followed by the addition of attention masks to give high weights to regions containing useful information.

\begin{figure}[t]
    \centering
    \includegraphics[width=\linewidth]{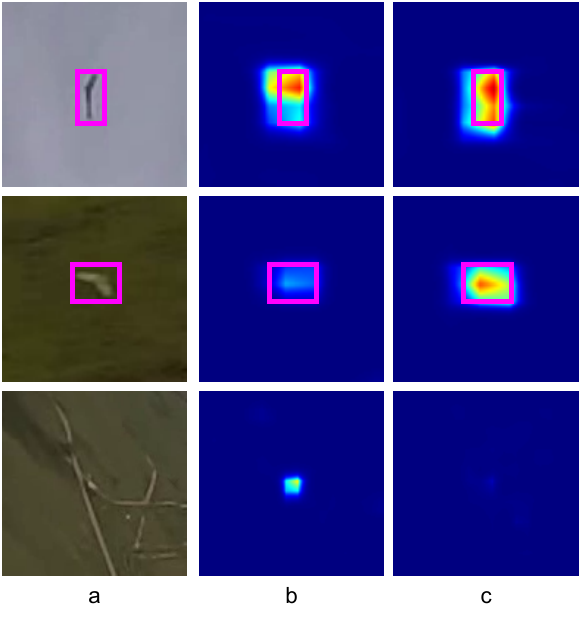}
    \caption{Role of attention. (a) Input image. (b) without and (c) with attention networks.   Top two rows shows the examples where attention networks helps the network to learn to give more weights to the pixels associated with drones and the last row represents the case where attention network suppress the non-drone pixels.} 
    \label{fig:attention_vs_non_attention}
\end{figure}

In the experiments, we observe that attention networks significantly help in achieving better drone localization. Note that the complete stage-1 is trained end-to-end where attention is learned automatically through network architecture, training data, and losses.  Figure \ref{fig:attention_vs_non_attention} demonstrates the difference in the network outputs which training with and without attention networks.

\subsection {Losses}
Drone detection datasets have two major challenges: There exists a large drone versus non-drone class imbalance i.e., the majority of pixels belong to the background and only a few pixels (if any) occupy the drone. Secondly,  due to extremely small size drones, the only difference of even 1 or 2 pixels between the detected box and ground truth brings down the IoU score to less than 0.5. Therefore, we use multiple losses to train our network.  Specifically, to address the class imbalance, focal loss \cite{lin2017focal} is used. and Distance-IOU loss \cite{Distance_IOU} is employed to achieve better IOU localization. Distance-IOU not only minimizes the IOU between the ground truth and detected bounding boxes but also reduces the distance between centers of two boxes. Finally, we use smooth-L1 loss \cite{ren2015faster} to jointly train pixel-wise attention network as shown in Figure \ref{fig:channel_pixel weighting_network} (b).

\subsection{Stage2: Exploiting Spatiotemporal Cues}

The purpose of this stage is to confirm true detections, reject the false detections and discover the missing detections of the stage-1. To find out the new probable drone locations, we use motion gradients which is explained below.\newline\newline
\noindent{\textbf{Motion boundaries:}} Drone can be characterized by locations undergoing motion.  However, since drone detection datasets involve moving camera, simple optical flow magnitude cannot be much useful. Therefore, we propose to use optical flow gradients to capture the change in motion. Specifically, given every three frames of a video, we first stabilize them using key-points detection and then compute forward and backward optical flow. After that maximum motion gradients across all three frames are computed as follows:
 
\begin{gather}
G= max(\sqrt{ u_{x}^2 + u_{y}^2}, \sqrt{v_{x}^2 + v_{y}^2 }), \\
M=max(G_{0\rightarrow1},G_{1\rightarrow2},G_{2\rightarrow1},G_{1\rightarrow0}),
\end{gather}
where $u_{x}$, $v_{x}$, $u_{y}$, and $v_{y}$  are optical flow gradients along x and y axis respectively, $M$ show motion boundaries and $G_{0\rightarrow1},G_{1\rightarrow2},G_{2\rightarrow1},G_{1\rightarrow0}$ represents motion gradients between frames, 0\textrightarrow1, 1\textrightarrow2, 2\textrightarrow1, 1\textrightarrow0 respectively. 

There are two limitations of motion boundaries:
The motion boundaries
provide high magnitude across the boundaries of drones and in most cases, do not completely cover drones. Secondly, due to the underlying approximation of optical
flow calculation,  usually, the maximum of optical flow gradient
magnitude does not match exactly with the moving drone. To address these issues, we dilate the motion boundaries and then apply conditional random field \cite{crf} to get the better localization of candidate drone regions. \newline
 
\noindent{\textbf{Cuboids formation:}} Given detections from stage-1 and newly discovered locations obtained using motion boundaries, our next step is to extract spatio-temporal features from all candidate drone locations. To this end, we initialize the correlation tracker at each candidate location (including stage-1 detections and newly discovered locations).  Due to small drone and complex camera motions, trajectories tend to drift away from their initial location within a few frames, therefore, we restrict trajectory length to eight frames. Specifically, given the candidate drone location, tracking is done three frames forward and four frames backward. Note that tracking is done after the motion stabilization of the corresponding eight frames. To capture contextual information across candidate location and to compensate for the trajectory drifting, N $\times$ N patches are extracted from video frames across each track, resulting in a cuboid of size N$\times$N$\times$8. Finally, to extract spatio-temporal features from each cuboid, we employ the Inflated-3D (I3D) network \cite{carreira2017quo}. We choose I3D due to its fast speed, small memory consumption, and excellent capability of capturing detailed spatio-temporal characteristics. To make the size of the cuboid consistent with that of the standard I3D network input dimensions, we use bilinear interpolation on each patch to resize the cuboid from N$\times$N$\times$8 to 224$\times$224$\times$8. 3D Convolution features are extracted from the third last layer of the I3D network which has dimensions of 14$\times$14$\times$480. To have consistency with the stage-1, we use bilinear interpolation to resize the feature maps to 60 $\times$60$\times$480. This is followed by 2D convolution layers to convert 60$\times$60$\times$480 to 60$\times$60$\times$2048 feature maps. Experimentally, we have also tried patch superresolution and feature map superresolution instead of resizing using bilinear interpolation, however, we did not observe any performance improvement. 

Finally, spatio-temporal convolution feature maps of each cuboid are aggregated across different scales using spatial pyramid pooling. This is followed by attention networks and network losses as discussed in Section 3.2

 \begin{figure}[t!]
    \centering
    \includegraphics[width=\linewidth]{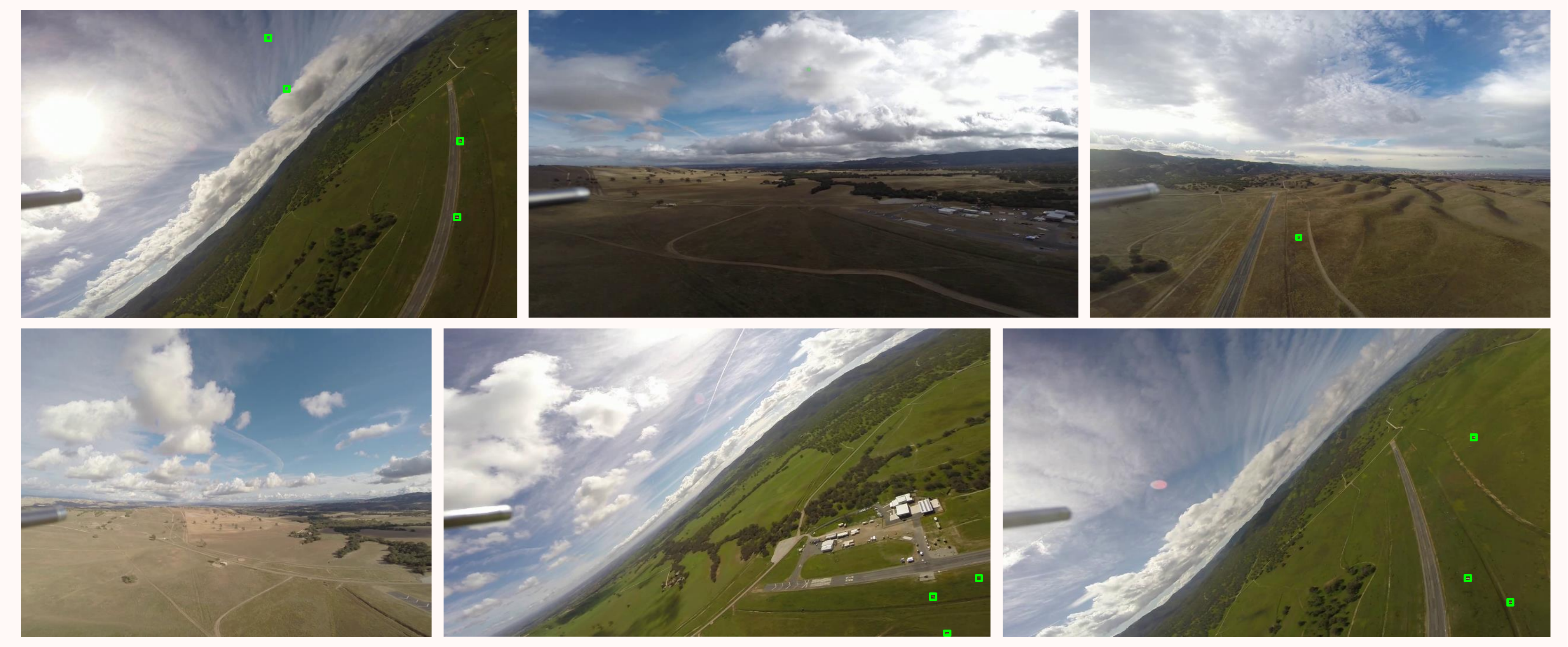}
    \caption{Samples frames from NPS-drone \cite{UAV_IROS} dataset. The green boxes enclose drones.}
    \label{fig:frame_samples_dataset1}
\end{figure}

\section{Experiments}

The main goal of our experiments is to verify that the proposed approach accurately localize drone in moving camera videos.
To this end, we perform extensive experiments on two challenging drone datasets and compare our approach with several competitive approaches, and analyze different components of the proposed approach.
\subsection{Implementation details} 
In this section, we provide implementation details of our approach.  During stage-1, we divide each frame into nine overlapping patches for NPS-Drones dataset \cite{UAV_IROS} and into four overlapping patches for FL-Drones dataset \cite{UAV_PAMI}. 
For both datasets, stage-1 is trained end to end from scratch, while for stage-2, a pre-trained I3D network with frozen weights is used for feature extractions.  
For the NPS-Drones dataset, a fixed 100$\times$100 patch size is extracted across the drone to make a cuboid while in the FL-Drones dataset size of the patch corresponds to the size of the drone.
In all the experiments, we use Adam \cite{kingma2014adam} optimizer with an initial learning rate of 0.001 without the decay parameter. 
 
A simple correlational tracker is employed during cuboid formation. We remove candidate locations (provided by motion boundaries), which are smaller or bigger than the maximum and minimum size of drones in the training data respectively. To improve training, we have also used hard-negative mining. In a post-processing step, we remove detections that only appear for a single frame.

\noindent\textbf{Evaluation metrics:} Following the related works, we evaluate the performance of our approach and baselines using precision, recall, F1-score, and average precision (AP) where every frame is treated
as an individual image for evaluation. We evaluate all the methods on every 4th frame of the testing data.

\begin{figure}[t]
    \centering
    \includegraphics[width=\linewidth]{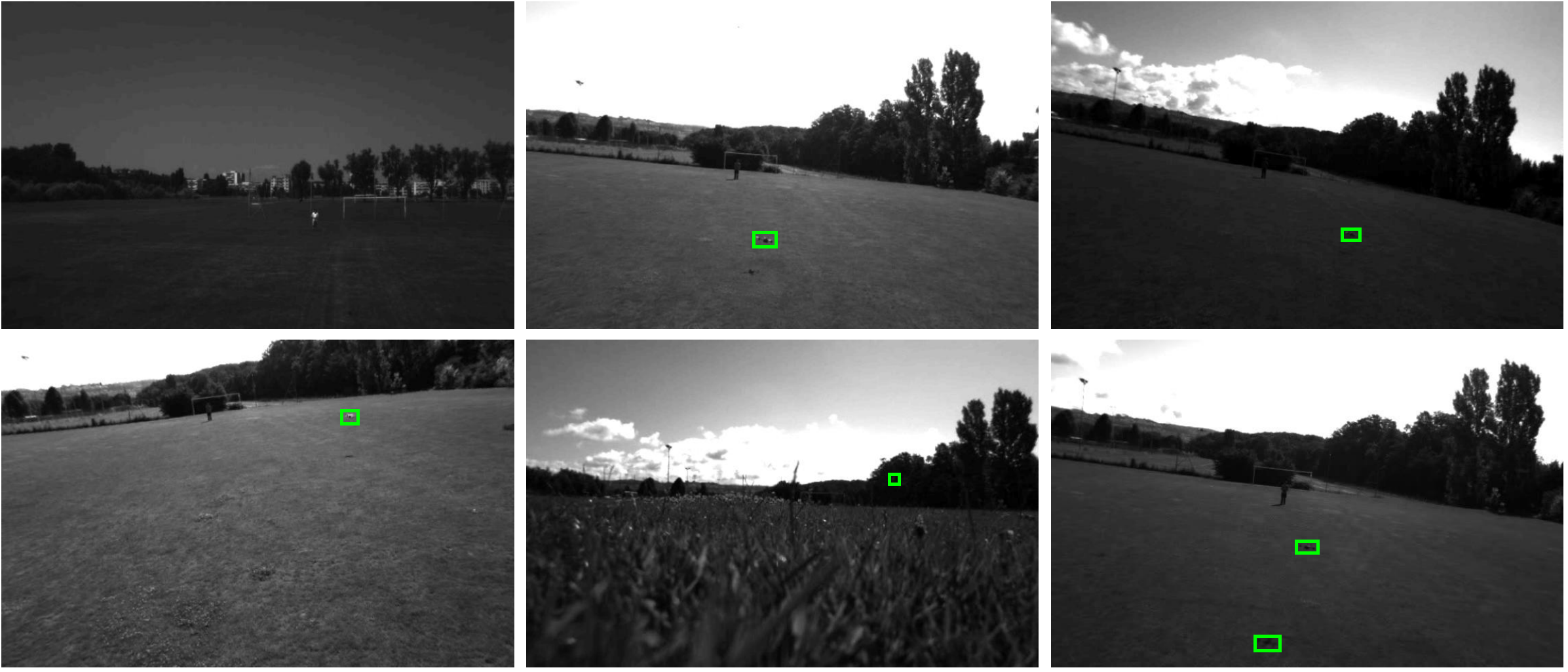}
    \caption{Samples frames from FL-drone \cite{UAV_PAMI} dataset. The green boxes enclose drones.}
    \label{fig:frame_samples_dataset2}
\end{figure}

\subsection{Datasets} 
We evaluate our approach on two drones datasets. Each of them is briefly introduced below.

\noindent\textbf{NPS-Drones \cite{UAV_IROS}:} This dataset is published by Naval Postgraduate School (NPS) and is publicly available\footnote{\url{https://engineering.purdue.edu/~bouman/UAV_Dataset/}}.  
The dataset contains 50 videos that were recorded at HD resolutions (1920$\times$1080 and 1280$\times$760) using GoPro-3 cameras mounted on a custom delta-wing airframe.  The minimum, average and maximum size of drones are 10 $\times$ 8, 16.2 $\times$ 11.6, and 65 $\times$ 21, respectively. 
The total number of frames in the dataset are 70250. In the experiments, the first 40 videos are used for training and validation and the last 10 videos are used for the testing.

\begin{figure}[h!]
    \centering
    \includegraphics[width=\linewidth]{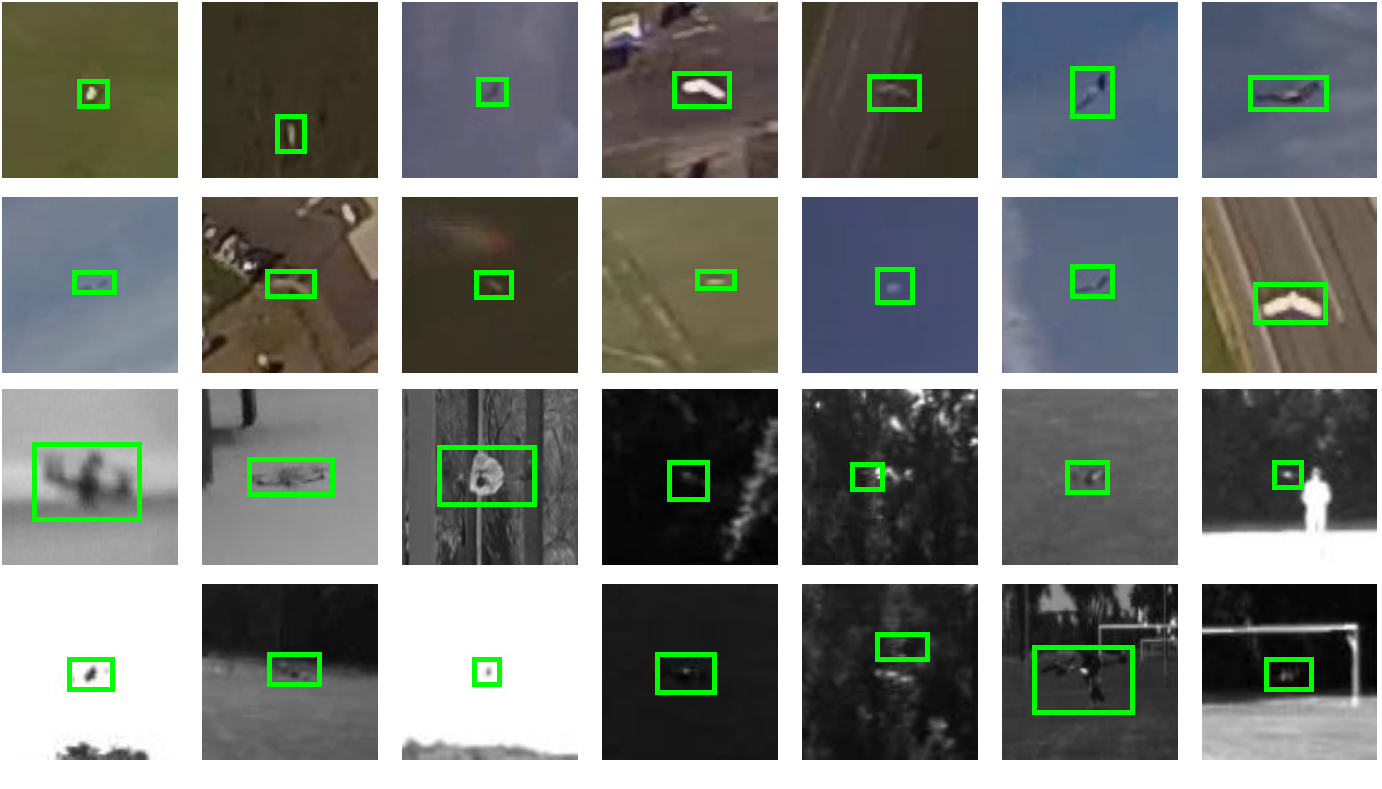}
    \caption{This figure shows the variability of drone shape and size in two datasets: NPS-drone \cite{UAV_IROS} (first two rows) and FL-drone \cite{UAV_PAMI} (last two rows) datasets. The green boxes present the ground truth bounding boxes.}
    \label{fig:uav_samples_dataset1}
\end{figure}

\noindent\textbf{FL-Drones \cite{UAV_PAMI}:} 
The second drone dataset used in the paper was introduced by Rozantsev et al. \cite{UAV_PAMI}. This dataset is quite challenging due to extreme illumination, pose, and size changes.
This dataset contains indoor and outdoor samples and the flying drones have variable shapes, and their shape is barely retained even in consecutive frames. Drones get mixed up in the background due to small size and intense lighting conditions coupled with complex background. The minimum, average, and maximum sizes of drones are: 9$\times$9, 25.5$\times$16.4, and 259$\times$197, respectively and frames resolutions are  640$\times$480 and 752$\times$480. This dataset contains 14 videos with a total of 38948 frames. As suggested by the authors \cite{UAV_PAMI}, half of the data is used for the training, and the other half is used for the testing.

\begin{figure}[ht]
    \centering
    \includegraphics[width=\linewidth]{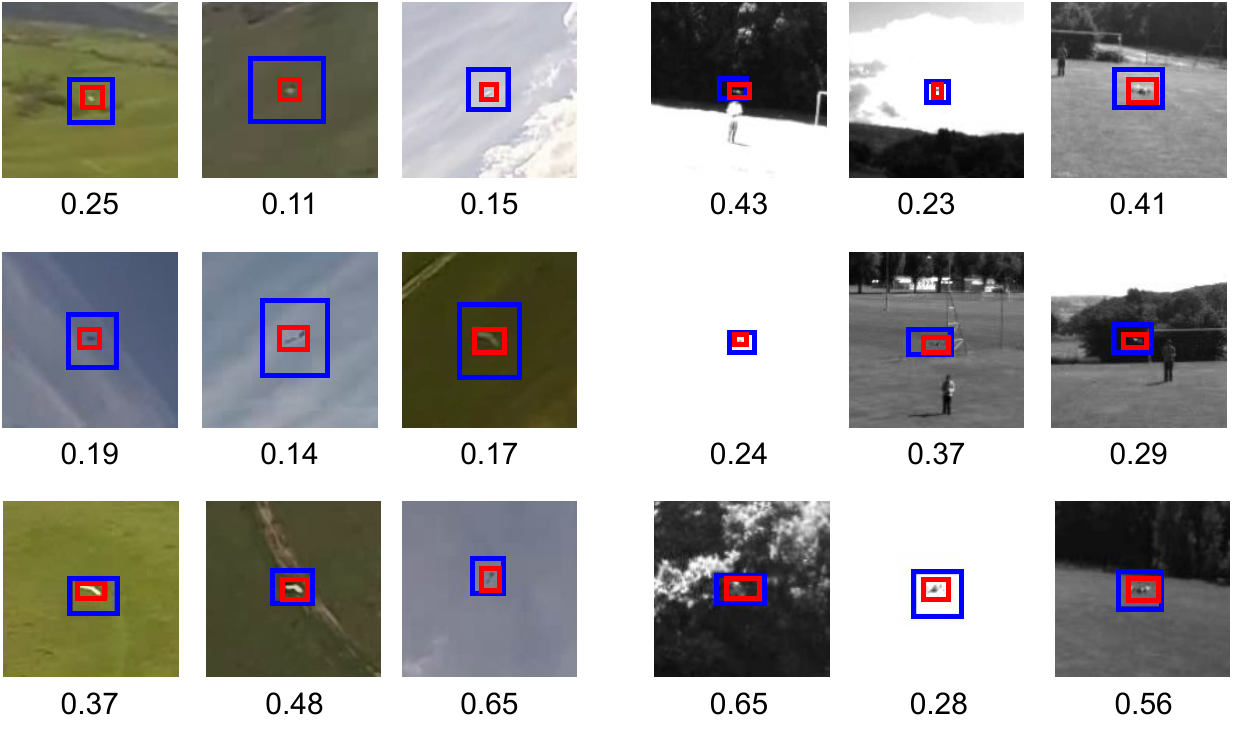}
    \caption{In the figure, we show the examples of our improved annotations. The blue boxes show the originally released annotations by the authors and the red boxes show our annotations done. The first three columns show the images from the NPS-drone dataset \cite{UAV_IROS} and the last three columns are images from the FL-drone dataset \cite{UAV_PAMI}. Under each image, we show the IOU between new and old annotations.}
    \label{fig:fixed_annotations}
\end{figure}

Sample frames and drones from both datasets are shown in Figure \ref{fig:frame_samples_dataset1}, Figure \ref{fig:frame_samples_dataset2} and, Figure \ref{fig:uav_samples_dataset1}. Note that there are several frames in both datasets without any drone. It is interesting to know that the majority of original annotations released by the authors of \cite{UAV_PAMI, UAV_IROS} are not precise and bounding boxes are much bigger than that of actual drones. To address this, we have re-annotated both datasets again. Given that number of frames in both datasets are more than 100K, re-annotation took significant time. The examples of improved annotations are shown in Figure \ref{fig:fixed_annotations}. These improved annotations will be released.

\begin{table}[h]
    \centering
    \begin{tabular}{| c | c | c | c | c |} 
        \hline
        \thead{Method} & \thead{Precision} & \thead{Recall} & \thead{F1 score} & \thead{AP} \\ [0.5ex]
        \hline\hline

        \makecell{SCRDet-H \cite{yang2019scrdet}} & 0.81 & 0.74 & 0.77  & 0.65  \\
        \hline

        \makecell{SCRDet-R \cite{yang2019scrdet}} & 0.79  & 0.71  & 0.75  & 0.61  \\
        \hline
        
        \makecell{FCOS \cite{tian2019fcos}} & 0.88 & 0.84 & 0.86 & 0.83 \\
        \hline
        
        \makecell{Mask-RCNN \cite{he2017mask}} & 0.66 & \textbf{0.91} & 0.76 & \textbf{0.89} \\
        \hline
        
        \makecell{MEGA \cite{chen2020memory}} & 0.88 & 0.82 & 0.85 & 0.83 \\
        \hline
        
        \makecell{SLSA \cite{wu2019sequence}} & 0.47 & 0.67 & 0.55 & 0.46 \\
        
        \hline
        \hline
        
        \makecell{Proposed} & \textbf{0.92} & \textbf{0.91} & \textbf{0.92} & \textbf{0.89} \\

        \hline
    \end{tabular}
    \caption{Quantitative comparison of the proposed approach with several state of the art approaches on the NPS dataset \cite{UAV_IROS}.}. 
    \label{tab:scores_table}
\end{table}

\begin{table}[h]
    \centering
    \begin{tabular}{| c | c | c | c | c |} 
        \hline
        \thead{Method} & \thead{Precision} & \thead{Recall} & \thead{F1 score} & \thead{AP} \\ [0.5ex]
        \hline\hline

        \makecell{SCRDet-H \cite{yang2019scrdet}} & 0.54 & 0.62 & 0.58 & 0.52  \\
        \hline
        
        \makecell{SCRDet-R \cite{yang2019scrdet}} & 0.55  & 0.62  & 0.58  & 0.52  \\
        \hline
        
        \makecell{FCOS \cite{tian2019fcos}} & 0.69 & 0.70 & 0.69 & 0.62 \\
        \hline
        
        \makecell{Mask-RCNN \cite{he2017mask}} & {0.76} & 0.68 & 0.72 & 0.68 \\ 
        \hline
        
        \makecell{MEGA \cite{chen2020memory}} & 0.71 & 0.72 & 0.71 & 0.65 \\
        \hline
        
        \makecell{SLSA \cite{wu2019sequence}} & 0.57 & 0.72 & 0.64 & 0.61 \\
        \hline
        \hline
        
        \makecell{Proposed} & \textbf{0.84} & \textbf{0.76} & \textbf{0.80} & \textbf{0.72} \\

        \hline
    \end{tabular}
    \caption{Quantitative comparison of the proposed approach with several state of the art approaches on the FL-Drones dataset \cite{UAV_PAMI}.}. 
    \label{tab:scores_table_2}
\end{table}
 
\subsection{Comparison with the State-of-the-art}
We compare the proposed approach with recently proposed approaches such as  fully convolutional one-stage object detector \cite{tian2019fcos}, cluttered and rotated small object detector \cite{yang2019scrdet}, instance segmentation \cite{he2017mask}, and video object detectors \cite{chen2020memory,wu2019sequence}. For comparison purposes, all the baseline
methods are finetuned using the pre-trained weights
available with public codes. Specifically,
SLSA \cite{wu2019sequence}, FCOS \cite{tian2019fcos}, MEGA \cite{chen2020memory}, and SCRDet \cite{yang2019scrdet} are finetuned
using corresponding papers’ pretrained weights and
MASK-RCNN \cite{he2017mask} is fine-tuned using
Resnet50 weights trained on ImageNet. I3D has the
weights obtained through training on Kinetics. Due to computational power limitations, all of the methods that we compared with are trained for iterations ranging from 80, 000 to 100, 000 using a single Nvidia 1080Ti GPU.

The quantitative comparison of proposed approach with the baselines over both drone detection datasets are shown in Table \ref{tab:scores_table} and Table \ref{tab:scores_table_2}.
The results demonstrate that our approach significantly outperforms the recent baselines on different evaluation metrics.

\subsection{Ablation studies}
\noindent{\textbf{Component analysis:}} We analyze different components of the proposed approach to verify their effectiveness. The experimental results in Table \ref{tab:ablation_study_table} enforce that each component of our approach is important and contributes toward final accuracy.
\begin{table}[h]
    \centering
    \begin{tabular}{| c | c | c | c | c |} 
        \hline
        \thead{Method} & \thead{Precision} & \thead{Recall} & \thead{F1 score} \\ [0.5ex]
        \hline\hline

        \makecell{CE} & 0.87 & 0.82 & 0.85  \\
        \hline
        
        \makecell{ $\mathcal{L}$} & 0.88 & 0.84 & 0.86  \\
        \hline
        
        \makecell{$\mathcal{L}$+ $\mathcal{CA}$} & 0.88 & 0.86 & 0.87  \\
        \hline
        
        \makecell{$\mathcal{L}$+ $\mathcal{PA}$} & 0.89 & 0.85 & 0.87  \\
        \hline
        
        \makecell{Stage-1} & {0.92} & {0.88} & {0.90} \\
        \hline
         \makecell{Stage-1+Stage-2} & \textbf{0.92} & \textbf{0.91} & \textbf{0.92} \\
        \hline
    \end{tabular}
    \caption{Ablation study of different components of our method on NPS dataset. The first row shows the proposed approach employing only cross-entropy loss (CE). The second row represents results incorporating focal loss and Distance-IOU loss ($\mathcal{L}$). The third row demonstrates results employing channel-wise attention ($\mathcal{L}$+ $\mathcal{CA}$). After that we show results using pixel-wise attention ($\mathcal{L}$+ $\mathcal{PA}$). Finally, the last two rows show results after stage-1 and stage-2 respectively.}
    \label{tab:ablation_study_table}
\end{table}

\begin{figure}[t]
    \centering
    \includegraphics[width=\linewidth]{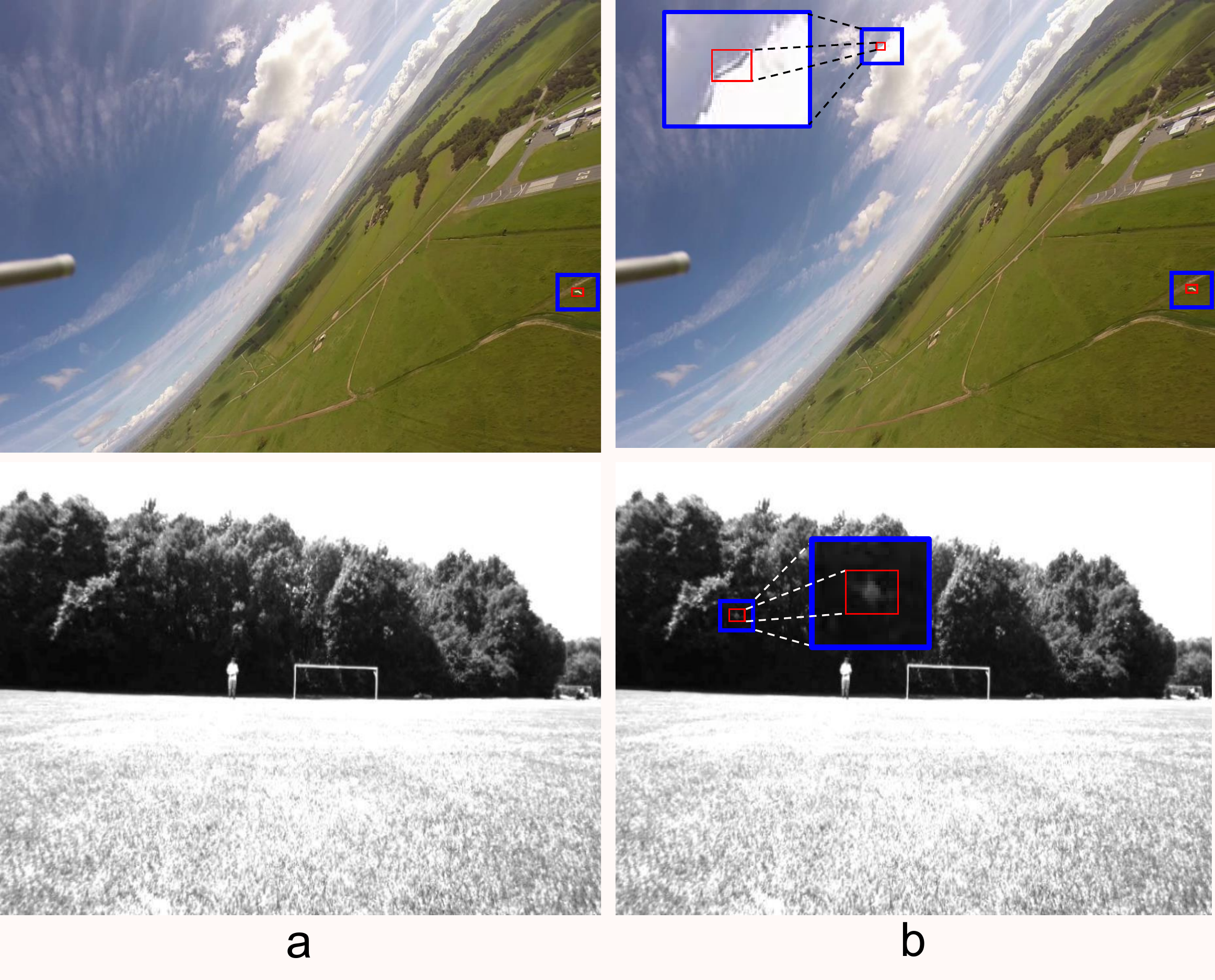}
    \caption{Qualitative comparison of one stage versus two stage detection results. (a) shows the detection results for stage-1 and (b) represents the detection results of two stage approach.  Red boxes represent detection and blues boxes are just for better visualizations. The first stage misses one drone in each example.}
    \label{fig:2d_vs_3d}
\end{figure}

\begin{figure}[t]
    \centering
    \includegraphics[width=\linewidth]{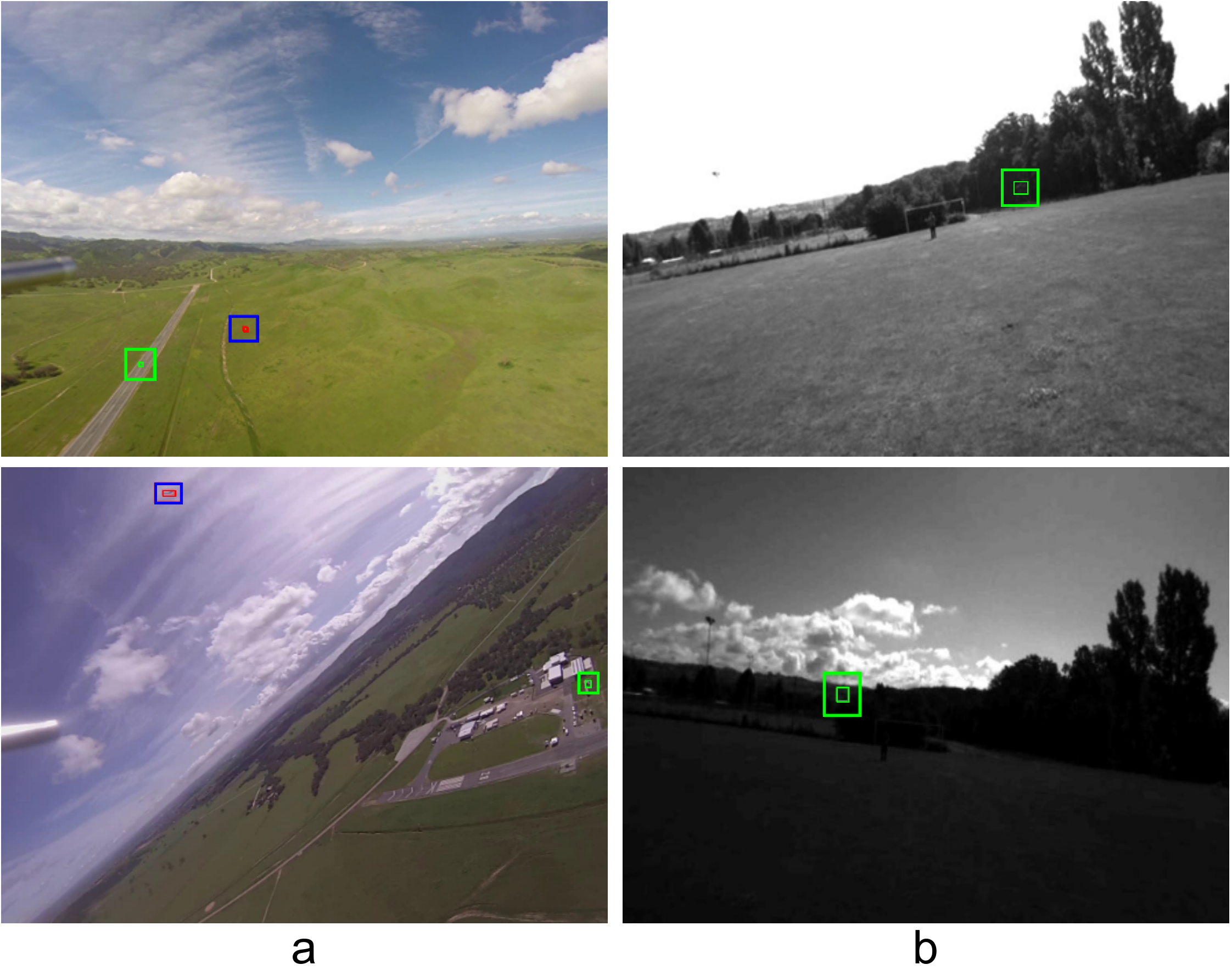}
    \caption{Failure cases of our method.
    Miss detections and correct detections are shown by green and red boxes respectively. Outer boxes are just for better visualizations.
    The drones with indistinguishable shape and motion are hard to detect.
    (a) and (b) represents samples from NPS-Drones \cite{UAV_IROS} and FL-Drones \cite{UAV_PAMI} datasets respectively.}
    \label{fig:failure_cases}
\end{figure}

\noindent{\textbf{Two Stage Approach:}} As shown in Table \ref{tab:ablation_study_table} and Fig \ref{fig:2d_vs_3d}, we observe that two-stage approach produce much better detection results as compared to 1-stage approach by discovering and classifying difficult drone locations. We have also tried multiple frames in 1-stage similar to \cite{CLusterNet} but achieved very low detection accuracy (below F1-score=40). We believe that due to the large and complex target and source drone motion, the 1-stage multiple frames approach is unable to learn to detect drones.\newline\newline
\noindent\textbf{Failure cases:} Since the proposed approach tries to detect drones based on appearance and motion cues, the drones with slow motion and indistinguishable shape are hard to detect. Figure \ref{fig:failure_cases} show failure cases of our approach.\newline

\section{Conclusion}
One of the most important challenges associated with drone use is collision avoidance or safe multi-drone flights. Therefore, it is crucial to develop robust computer vision methods that can detect and enable collision avoidance using inexpensive cameras. We have presented a two-stage approach for drone detection from other flying drones employing spatio-temporal cues. Instead of relying on region proposal-based methods, we have used a segmentation-based approach for accurate drone detection using pixel and channel-wise attention. In addition to using appearance information, we have also exploited motion information between frames to get a better recall. We observe that for drones to drones videos, the two-stage approach performs better than the one-stage approach.
Thorough comparisons with the state of the arts and detailed ablation studies validate the framework and the ideas proposed in this work.
\newline\newline\noindent{\textbf{Acknowledgements:}}
We would like to thank Mohsen Ali and Sarfaraz Hussein for their insightful discussions.

{\small
\bibliographystyle{ieee}
\bibliography{ms}
}

\end{document}